%% file: iclr2026_conference.tex
\documentclass{article} 
\usepackage{iclr2026_conference,times}

\input{math_commands.tex}

\usepackage{hyperref}
\usepackage{url}
\usepackage{booktabs}
\usepackage{algorithm}
\usepackage{algorithmic}
\usepackage{xcolor}
\usepackage{amsmath}
\usepackage{tabularray}
\usepackage{subcaption}

\usepackage{newfloat}
\usepackage{listings}

\usepackage{times}  
\usepackage{helvet}  
\usepackage{courier}  
\usepackage{url}  
\usepackage{graphicx} 

\usepackage{listings}
\usepackage{xcolor}

\definecolor{codegreen}{rgb}{0,0.6,0}
\definecolor{codegray}{rgb}{0.5,0.5,0.5}
\definecolor{codepurple}{rgb}{0.58,0,0.82}
\definecolor{backcolour}{rgb}{0.95,0.95,0.92}

\lstdefinestyle{pythonstyle}{
    backgroundcolor=\color{backcolour},   
    commentstyle=\color{codegreen},
    keywordstyle=\color{magenta},
    numberstyle=\tiny\color{codegray},
    stringstyle=\color{codepurple},
    basicstyle=\ttfamily\footnotesize,
    breakatwhitespace=false,         
    breaklines=true,                 
    captionpos=b,                    
    keepspaces=true,                 
    numbers=left,                    
    numbersep=5pt,                  
    showspaces=false,                
    showstringspaces=false,
    showtabs=false,                  
    tabsize=2,
    language=Python
}

\iclrfinalcopy

\newcommand{\name}{\textsc{Mortar}}
\newcommand{\names}{\textsc{Mortar's}}

\title{Mortar: Evolving Mechanics for \\ Automatic Game Design}

\author{   Muhammad U. Nasir$^{1,2}$ \quad Yuchen Li$^2$ \quad Steven James$^1$ \quad Julian Togelius$^2$\\
$^1$University of the Witwatersrand \quad $^2$New York University\\
\texttt{\{muhammad.nasir,steven.james\}@wits.ac.za} \quad 
\texttt{yl6394@nyu.edu} \\
\texttt{julian@togelius.com}
}

\begin{document}

\maketitle

\begin{abstract}
We present \name{}, a system for autonomously evolving game mechanics for automatic game design. Game mechanics define the rules and interactions that govern gameplay, and designing them manually is a time-consuming and expert-driven process. \name{} combines a quality-diversity algorithm with a large language model to explore a diverse set of mechanics, which are evaluated by synthesising complete games that incorporate both evolved mechanics and those drawn from an archive.
The mechanics are evaluated by composing complete games through a tree search procedure, where the resulting games are evaluated by their ability to preserve a skill-based ordering over players---that is, whether stronger players consistently outperform weaker ones. We assess the mechanics based on their contribution towards the skill-based ordering score in the game. We demonstrate that \name{} produces games that appear diverse and playable, and mechanics that contribute more towards the skill-based ordering score in the game. We perform ablation studies to assess the role of each system component and a user study to evaluate the games based on human feedback.
\end{abstract}


\section{Introduction}

Procedural content generation (PCG) is a well-studied approach in game design, concerned with the automatic creation of game content such as levels, maps, items and narratives \citep{shaker2016procedural,liu2021deep}.
PCG serves multiple purposes: enabling runtime content generation in games such as roguelikes, providing ideation tools for designers, automating the production of repetitive content, and facilitating research into creativity and design processes.
Traditionally, PCG research has focused on structural aspects of games---particularly level or layout generation \citep{risi2020increasing}---where the goal is to produce environments that are coherent, solvable, and varied. 

By contrast, comparatively little attention has been paid to the procedural generation of \emph{game mechanics}---the underlying rules for interactions that govern gameplay. Yet mechanics play a central role in shaping the player experience, determining not just how players act, but what kinds of strategies and emergent behaviours are possible. Designing mechanics is inherently challenging: unlike levels, which can be evaluated by solvability or novelty, the utility of a mechanic depends on the dynamics it induces within the context of a game. This makes both generation and evaluation significantly harder.

A central premise of this work is that evaluating game mechanics is fundamentally more difficult than evaluating assets or level layouts. Unlike these forms of content, a mechanic \emph{cannot be judged in isolation}---it only gains meaning through the gameplay it enables. A mechanic that appears novel or complex may still be uninteresting if it does not support skill-based interaction. This insight motivates our approach: effective automation of mechanic design requires not only a generative model, but also a principled way to assess a mechanic's utility in the context of play.

We address this challenge by introducing a mechanic-centric framework for automatic game design. The central idea is to evolve mechanics not in isolation, but through their contribution to the quality of full games. Specifically, we evaluate mechanics by constructing complete games around them, and measuring whether the resulting games induce a skill-based ordering over players of different capabilities. This allows us to define a concrete notion of usefulness for a mechanic: its contribution to the overall expressivity and skill gradient of the games in which it appears.

We introduce \name{}, a system that evolves game mechanics using a quality-diversity algorithm guided by a large language model (LLM). 
\name{} maintains a diverse archive of mechanics, represented as code snippets, which are mutated and recombined through LLM-driven evolutionary operators.
Each evolved mechanic is evaluated by embedding it into full games constructed via Monte Carlo Tree Search, which incrementally builds games by composing mechanics from the archive. 
These games are evaluated based on their ability to induce a consistent skill-based ranking over a fixed set of agents.
We define a novel fitness measure, which quantifies the contribution of a mechanic to the final game's skill-based ordering, inspired by Shapley values \citep{shapley1953value}.

We demonstrate that \name{} can evolve a diverse set of game mechanics that contribute to the quality and playability of the generated games.\footnote{Play the generated games at: \url{https://mortar-x3p7.onrender.com/}} The resulting games exhibit coherent structure, varied interaction patterns, and meaningful skill gradients. Through ablations, we show that both the tree-search-based composition and the LLM-driven mutations are critical for generating high-quality mechanics. Our results highlight the potential for using LLMs not only as generators, but as evaluators and collaborators in the game design loop. Compared to existing work on game generation, \name{} is the first system to apply an LLM-driven quality-diversity method to video game generation, which is a significantly different problem than board game generation; we also show that it is possible to do this in code space without a domain-specific language, and introduce a novel fitness measure.

The system described here is a research prototype for the purposes of understanding how to best generate complementary game mechanics. However, it could also serve as an ideation tool for game designers, suggesting new mechanics and mechanic combinations, perhaps in response to designer input. It is not meant to generate complete games, and aims to empower rather than replace game designers.

\begin{figure*}[t]
\centering
\includegraphics[width=1\textwidth]{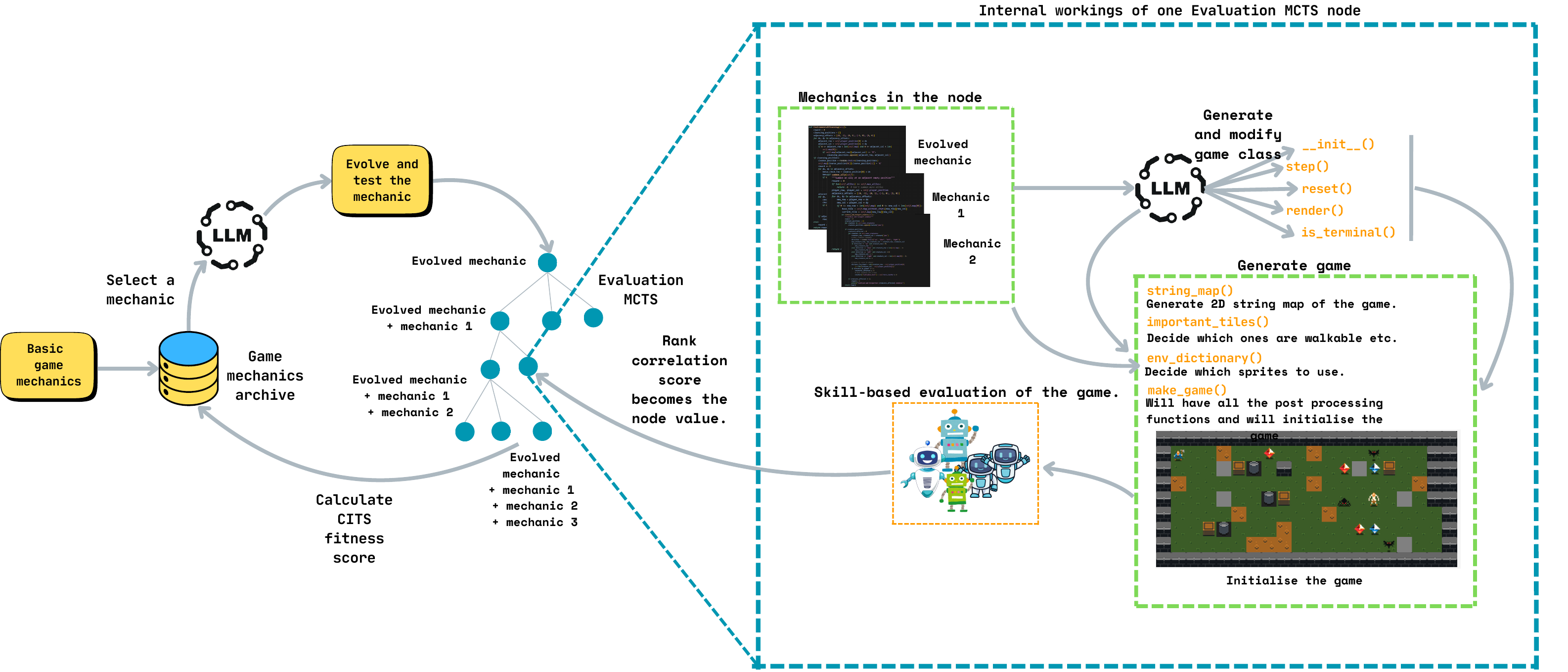} 
\caption{A flow diagram of \name{}}
\label{flowfig}
\end{figure*}

\section{Method}

\name{} is an evolutionary algorithm for generating game mechanics, where a large language model (LLM) is used to implement code-level variation operators. A core principle of the method is that a mechanic's value lies in the gameplay it enables; mechanics are evaluated not in isolation, but by the contribution they make to full games. We formalise this through a notion of \textit{importance}, which guides the search process.

\subsection{Evolution Setup}

\name{} employs a Quality-Diversity (QD) algorithm, using a fixed 2D archive (as in MAP-Elites~\citep{mouret2015illuminating}) to store and explore diverse game mechanics. We refer to this structure as the \textit{Mechanics Archive}. Each mechanic is represented as a Python function belonging to a game class, and placed in the archive based on two behavioural descriptors:

\begin{enumerate}
    \item \textbf{Mechanic Type}: A categorical descriptor indicating the gameplay behaviour the mechanic enables. We define 8 mechanic types (detailed in Section~\ref{sec:exp}), each associated with 10 descriptive category words. To classify an evolved mechanic, we compute similarity scores between the mechanic's name and all category words, creating a normalised similarity vector. The mechanic type is determined by identifying the highest similarity score's index and multiplying this index by the score to produce a positional similarity value that serves as the behavioural descriptor.

    \item \textbf{Code Complexity}: Computed using weighted Abstract Syntax Tree (AST) analysis. We parse the mechanic's code into an AST representation and calculate complexity as a weighted sum of function calls, assignments, and return statements. Function calls receive the highest weight, as mechanisms requiring more function calls exhibit greater complexity. Assignments are weighted to reflect that additional variables may enable more interesting behaviours. Return statements contribute to complexity scoring because multiple exit paths can produce diverse behavioural outcomes.

\end{enumerate}


Mechanics are selected from the archive and modified using several LLM-implemented evolutionary operators: \textit{Mutation} adds new functionality to a single mechanic; \textit{diversity mutation} samples three mechanics and prompts the LLM to generate a behaviorally distinct variant; \textit{crossover} merges two mechanics (selected based on AST similarity) into a functional combination that integrates elements from both; and \textit{compatibility mutation} generates mechanics that complement existing ones in a game, primarily used during game evaluation (see subsequent sections).

\subsection{Evaluating Game Mechanics}

Each evolved mechanic is represented as a function within a Python class. To prepare it for evaluation, we prompt the LLM to construct the rest of the class around it in a step-by-step fashion, starting with the \texttt{\_\_init\_\_()} method to define any required variables and scaffolding, \texttt{step} method to add actions, \texttt{reset} and \texttt{render} method as needed. 

The mechanic is then tested for syntax and runtime errors. If no errors occur, we simulate it in a static test environment with simple objects and characters for the mechanic to interact with them, if necessary. A Monte Carlo Tree Search (MCTS) agent is used to interact with the environment, verifying that the mechanic is functional and non-trivial. Only mechanics that pass both tests proceed to the \textit{usefulness} evaluation stage. Failed mechanics are discarded to reduce unnecessary LLM calls.


\subsection{Automated Game Construction}

To evaluate a mechanic’s usefulness, we embed it within a full game. Games are constructed through MCTS, where the root node is the evolved mechanic, and each expansion adds a new mechanic that is either sampled from the archive or generated via compatibility mutation. Each path through the tree represents a particular combination of mechanics; that is, a complete game.

Games are also implemented as Python classes, following a common template with core methods, such as \texttt{step}, \texttt{reset}, \texttt{render}, move mechanics and preset variables. The LLM is prompted to modify or add functionality to these methods as needed, in an iterative manner. It is also asked to define any helper methods or variables required by the mechanics.
At the end of this process, the LLM is prompted to define a win condition and generates a corresponding termination function. It also selects appropriate tiles from a predefined set, maps them to characters, and generates a 2D string-based level layout using these mappings. A final function defines which tiles are walkable, interactive, or character-specific.

The complete game script includes the game class, 
the tile and map generation functions, and a preset function that instantiates the full game.
Simple postprocessing ensures the map is rectangular, contains exactly one player, and is free of formatting issues (e.g. whitespace padding).

\subsection{Evaluation of the Game}

A central idea in our work is that a game's quality is revealed through the emergence of a consistent skill gradient or game depth: a well-designed game should allow players of differing abilities to be meaningfully distinguished.
We implement this by evaluating how well the game ranks a fixed set of players by skill. This approach allows us to evaluate not just whether a game is playable, but whether it rewards skill---a more robust signal of design quality.

To assess whether a game rewards skill, we fix a pool of five agents with varying ability levels, inspired by~\cite{nielsen2015general}. These include three MCTS agents with increasing numbers of rollouts, a random agent, and an agent that takes no actions.\footnote{Any agents with a clear capability ordering would suffice, such as heuristic agents with different search depths, or reinforcement learning agents with varying training budgets.} This defines a clear expected skill ordering: the strongest agent should be the MCTS variant with the most rollouts, followed by the medium and low rollout agents, then the random agent, and finally the no-op agent.
The outcome rank is induced by playing the game and recording empirical win rates. To quantify alignment between the expected and outcome rankings, we compute Kendall's Tau ($\tau$), a standard measure of rank correlation: $\tau = \frac{C - D}{\frac{p(p - 1)}{2}}$. Here, $C$ and $D$ are the number of concordant and discordant pairs, respectively, and $p$ is the number of players (five in this case). Concordance occurs when the relative ranking between two players agrees between the expected and observed orders; discordance occurs when they disagree. A value of $1$ indicates perfect alignment with the expected ranking, $0$ indicates no correlation, and $-1$ reflects a completely reversed order. We consider a game unplayable if $\tau = -1$.


While $\tau$ provides a global measure of game quality, it reveals nothing about the source of that quality. To address this, we introduce \textit{Constrained Importance Through Search} (CITS), a scoring method to measure each mechanic's marginal contribution to the emergence of a skill gradient.
Inspired by Shapley values \cite{shapley1953value}, CITS estimates how much each mechanic contributed to the final game's $\tau$ score. However, computing full Shapley values would require evaluating every subset of mechanics--exponential in the number of mechanics. Instead, CITS is defined over the exploration tree constructed during generation, making it computationally tractable and grounded in actual gameplay evaluations.
Formally, the CITS score for mechanic $i$ is:

\begin{equation*}
\mathrm{CITS}_i = \frac{1}{|N_i|} \sum_{n \in N_i} \phi_i^{(n)}, 
\end{equation*}

where $N_i = \{n \in T : i \in M_n, n \neq n_{root}\}$ is the set of non-root nodes in the tree $T$ that contain mechanic $i$, and $M_n$ is the mechanic set at node $n$. 
The contribution $\phi_i^{(n)}$ is computed using the standard Shapley formula over the restricted set of explored subsets:


\begin{equation*}
\phi_i^{(n)} = \sum_{S \subseteq M_n \setminus \{i\}} \frac{|S|! \cdot (|M_n| - |S| - 1)!}{|M_n|!} \cdot \Delta_i^{(n)}(S),
\end{equation*}
where the marginal value term is defined as the difference in value when adding mechanic \(i\) to the subset \(S\):

\begin{equation*}
\Delta_i^{(n)}(S) = v_T(S \cup \{i\}) - v_T(S).
\end{equation*}

Finally, the value function $v_T(S)$ returns the $\tau_m$ score for the node $m$ with exactly mechanics $S$, if such a node exists in the tree; otherwise, it is defined to be $0$:

\begin{equation*}
    v_T(S) = \begin{cases}
\tau_m & \text{if } \exists m \in T \text{ s.t. } M_m = S \\
0 & \text{otherwise}
\end{cases}
\end{equation*}

This search-constrained Shapley approach allows us to assess a mechanic's value in context, measuring its contribution within actual, discovered game designs rather than hypothetical combinations. As such, the CITS score provides a principled, interpretable, and efficient mechanism for attributing gameplay quality to individual mechanics.



\section{Experiment Setup} \label{sec:exp}

\name{} employs a 2D Quality-Diversity (QD) archive with dimensions for mechanic type (0--1.0) and code complexity (4--40), forming a 13 $\times$ 13 grid. The first dimension categorises mechanics into nine types: \textit{movement, interaction, combat, progression, environment, puzzle, resource management, exploration, time manipulation}.
To categorise a mechanic, we use  DistilBERT~\citep{sanh2019distilbert} embeddings to compute the similarity between mechanic function names and associated category words (detailed in the Appendix~\ref{appendixC}). The complexity dimension and archive ranges were determined through experimentation to maximise archive coverage.

The system operates with a batch size of 10, selecting individuals from the archive and applying evolutionary operators in parallel. Operator selection probabilities are 50\% for diversity mutation, 30\% for mutation, and 20\% for crossover. Diversity mutation samples three mechanics, while crossover selects pairs based on AST similarity. The static environment used to evaluate the evolved mechanic in isolation can be found in the Appendix~\ref{appendixA}. 

\begin{figure*}[t]
\centering
\includegraphics[width=1\textwidth]{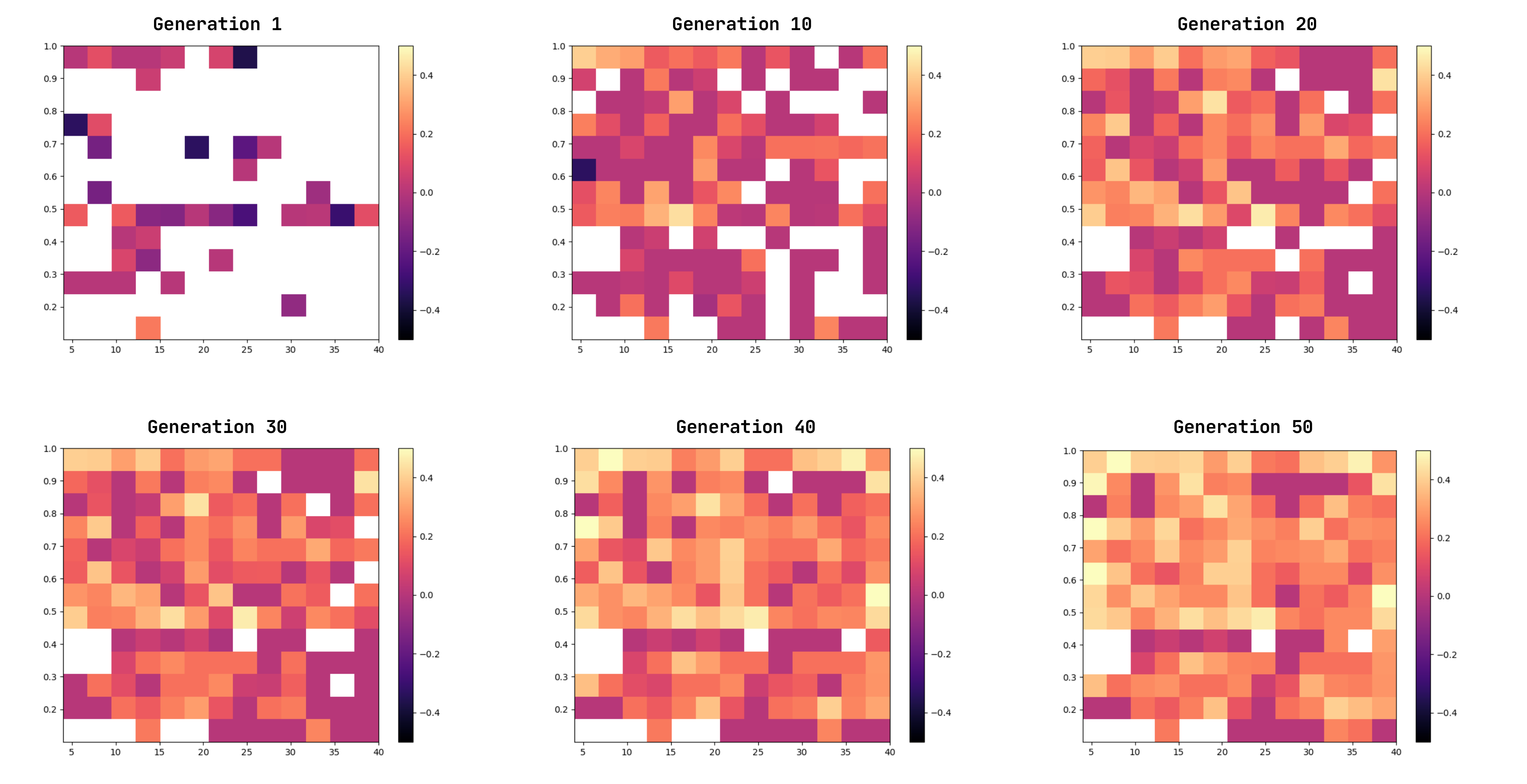} 
\caption{Coverage of game mechanic archive over a run.}
\label{coveragefig}
\end{figure*}

For game construction, we use MCTS with 20 iterations, where each expansion adds one mechanic per node (maximum 3 children per node). Unlike traditional MCTS, we do not simulate; instead, we evaluate the complete game formed by all mechanics on the path from root to the newly expanded node, then backpropagate before proceeding to the next expansion. Compatibility mutation generates new mechanics within nodes, with a 50\% probability of creating novel mechanics versus selecting from the existing archive. All LLM operations use GPT-4o-mini for both evolution and game creation. Skill assessment employs five agents with a clear capability ordering: MCTS variants with 100,000, 10,000, and 1,000 iterations, plus random and no-action agents for Kendall's Tau rank correlation computation.

We conduct extensive ablation studies, replacing the MCTS procedure with three alternatives: random mechanic selection, LLM-prompted selection, and greedy fitness-based selection. Each method generates games with 1-4 mechanics to compute CITS scores. We then conduct another ablation with a \textit{Sokoban}~\citep{murase1996automatic} level as the initial game. All experiments are averaged over five runs due to computational constraints (approximately \$30--50 per run with GPT-4o-mini).

Our evaluation metrics assess \name{}'s progression through multiple measures. The Quality-Diversity (QD) Score sums all fitness values to indicate improving mechanic quality. Accumulated Rank Correlation totals Kendall's $\tau$ scores across all tree nodes. We track both maximum and mean fitness scores via CITS evaluation, monitor the number of elites filling the archive, and calculate game creation success rate as the proportion of functional games among all generated games.

Furthermore, a user study is conducted to get feedback to known if the games are actually interesting or not. We provide 6 generated games, and pair them together according to their distribution. We then ask the user to play the games and mark which one of the two is more \textit{interesting}, \textit{novel}, \textit{fun to play}, and \textit{easy to understand}. We also give them an option of \textit{Neither}, which is very important to us as it will let us know if the games are actually meaningful. 

\section{Results}

\begin{figure}[h]
\centering
\begin{subfigure}[b]{0.48\textwidth}
    \centering
    \includegraphics[width=\textwidth]{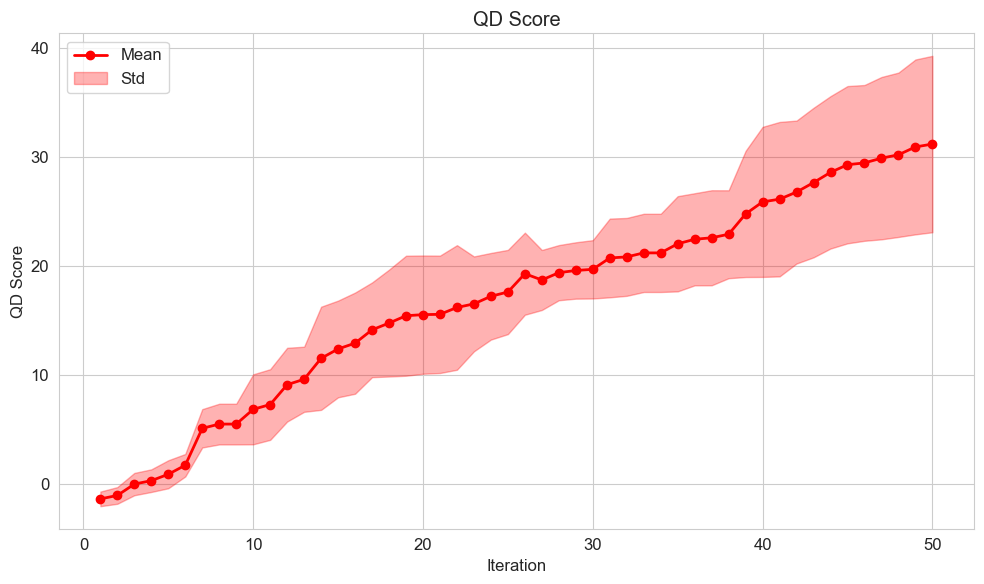}
    \caption{QD score}
    \label{qdf}
\end{subfigure}
\hfill
\begin{subfigure}[b]{0.48\textwidth}
    \centering
    \includegraphics[width=\textwidth]{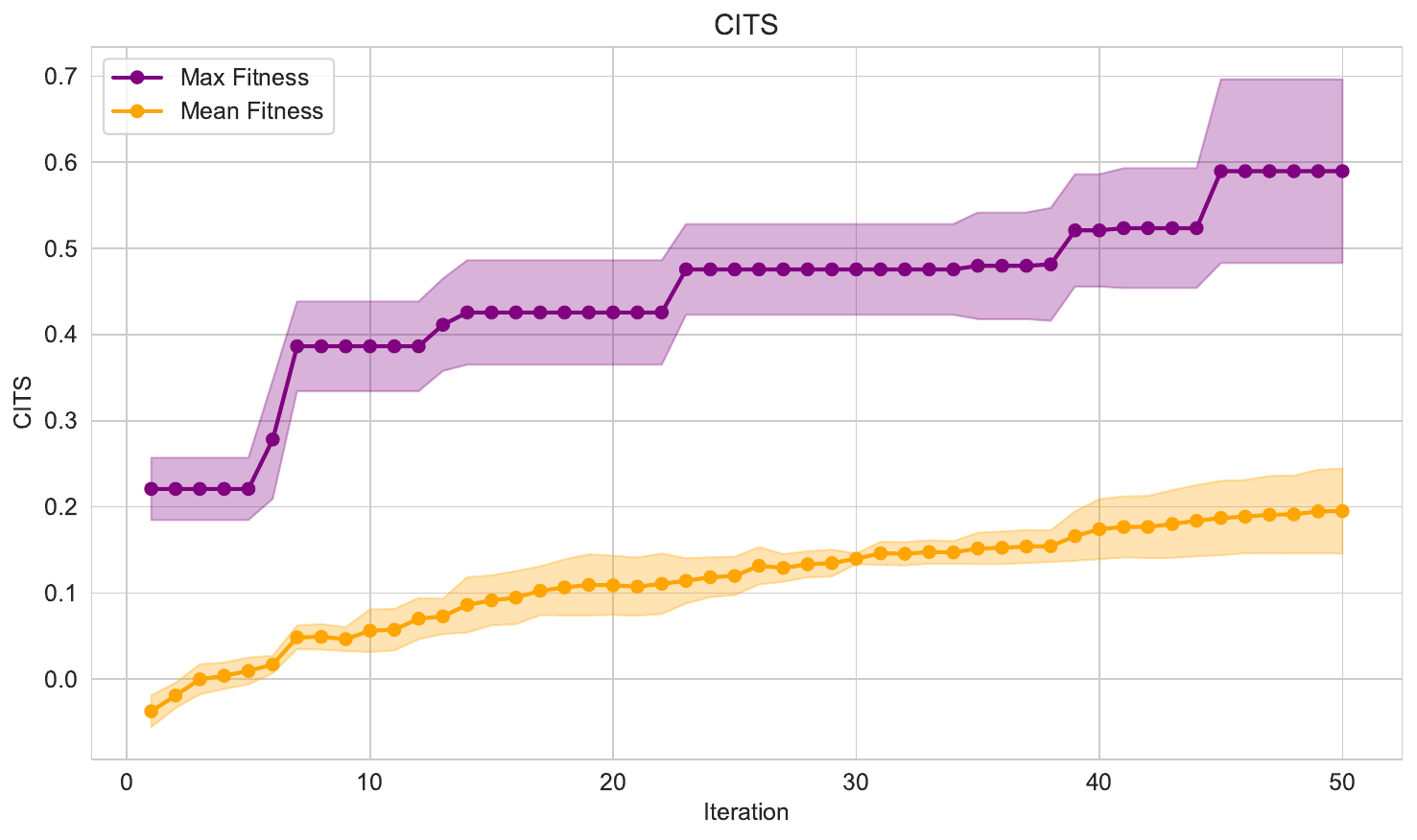}
    \caption{Mean and maximum CITS score, representing fitness of a game mechanic in the archive.}
    \label{fitnessfig}
\end{subfigure}

\begin{subfigure}[b]{0.6\textwidth}
    \centering
    \includegraphics[width=\textwidth]{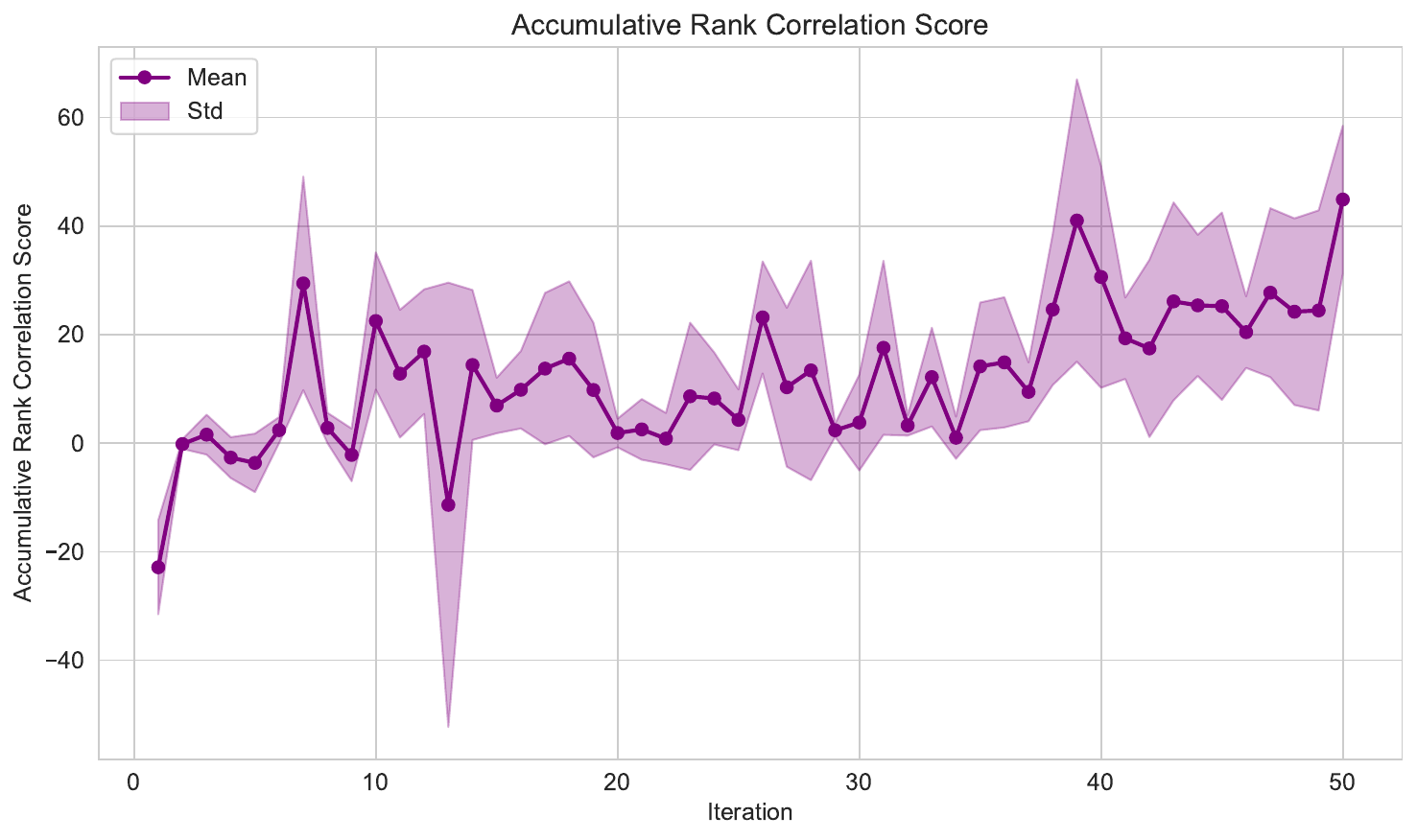}
    \caption{Accumulative Kendall's $\tau$ rank correlation score.}
    \label{rootfig}
\end{subfigure}

\caption{Performance metrics over evolutionary generations.}
\label{fig:combined}
\end{figure}

In this section, we analyse results from the complete \name{} pipeline, ablation and user studies. The QD score, which sums fitnesses of all archive individuals, demonstrates \name{}'s ability to evolve increasingly better mechanics over time (Figure~\ref{qdf}). Figure~\ref{fitnessfig} reveals complementary patterns: mean fitness (CITS score) increases gradually across generations while maximum fitness shows stepwise improvements, indicating \name{}'s capacity for continued mechanic discovery. Figure~\ref{rootfig} provides additional evidence of progression through the accumulative Kendall's $\tau$ rank correlation score per Evaluation MCTS tree, showing that \name{} increasingly identifies engaging games that exhibit meaningful skill-based player rankings across generations.


Table~\ref{tab:comparison} compares \name{} with alternative approaches to the MCTS evaluation component, our core methodological contribution. \name{} demonstrates superior evolvability through the highest archive coverage (Figure~\ref{coveragefig}) and consistently achieves the best QD score, maximum CITS score, and mean CITS score, indicating its ability to discover higher-quality mechanics. The results are shown in Table~\ref{tab:comparison}. While Greedy Search achieves a marginally better game creation success rate---likely because it always selects the most fit mechanics---this suggests that highly fit mechanics have greater potential for generating playable games. However, \name{}'s comprehensive performance across multiple metrics demonstrates the effectiveness of its tree search-based composition approach for mechanic evolution. Furthermore, \textit{Sokoban Initialisation} suggests that the \name{} is sensitive to the initial mechanics and game layout, which impacts evolvability, as evidenced by the very low number of elites in this case.

\begin{table*}[h]
\centering
\caption{Comparison of \name{} with alternative mechanic selection strategies: LLM-based selection, random selection, and greedy fitness-based selection across quality-diversity metrics.}
\resizebox{\textwidth}{!}{%
\begin{tabular}{lccccc}
\toprule
Method & No. of elites~$\uparrow$ & QD score~$\uparrow$ & Max CITS~$\uparrow$ & Mean CITS~$\uparrow$ & Games success rate~$\uparrow$ \\
\midrule
Evaluation MCTS (ours) & \textbf{155}~$\pm$~\textbf{4.51} & \textbf{31.18}~$\pm$~\textbf{8.10} & \textbf{0.59}~$\pm$~\textbf{0.11} & \textbf{0.20}~$\pm$~\textbf{0.05} & 16.97~$\pm$~4.64 \\
LLM Selection    & 141~$\pm$~5.83 & 17.64~$\pm$~4.91 & 0.27~$\pm$~0.09 & 0.13~$\pm$~0.06 & 11.69~$\pm$~5.14 \\
Random Selection & 144~$\pm$~9.10 & 9.86~$\pm$~6.71 & 0.14~$\pm$~0.08 & 0.06~$\pm$~0.04 & 11.77~$\pm$~4.19 \\
Greedy Selection & 139~$\pm$~5.15 & 25.37~$\pm$~2.81 & 0.51~$\pm$~0.06 & 0.18~$\pm$~0.13 & \textbf{18.24}~$\pm$~\textbf{1.19} \\
Sokoban Initialisation & 110~$\pm$~10.52 & 15.19~$\pm$~3.12 & 0.45~$\pm$~0.12 & 0.19~$\pm$~0.07 & 15.11~$\pm$~2.83 \\
\bottomrule
\end{tabular}}
\label{tab:comparison}
\end{table*}

\begin{figure*}[h!]
\centering
\includegraphics[width=0.9\textwidth]{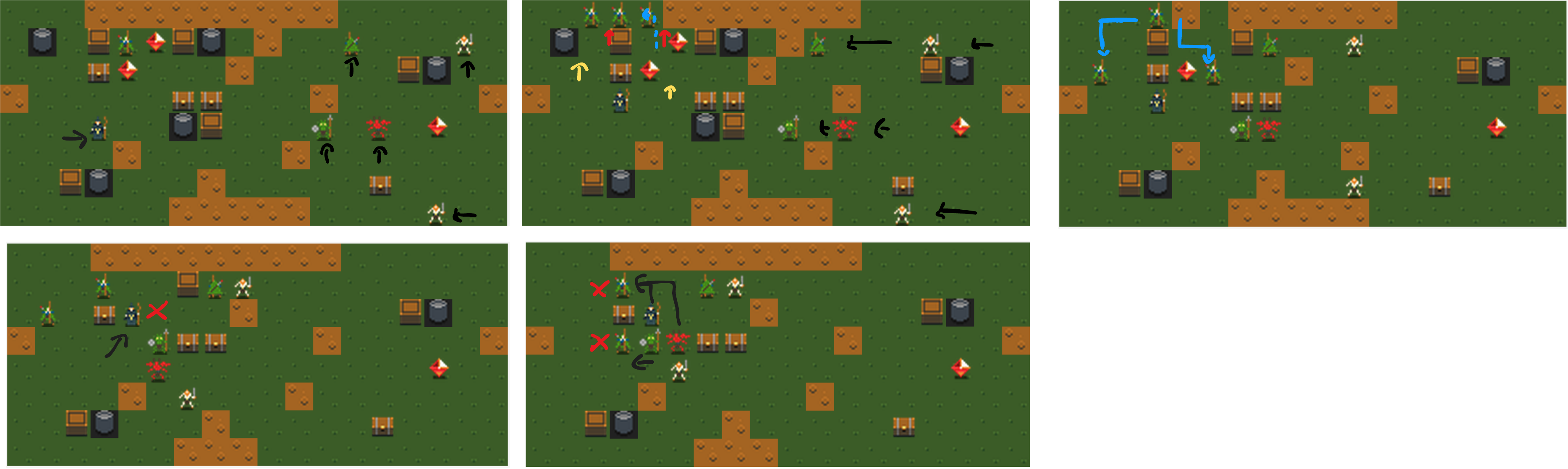} 
\caption{\textit{AllyCraft} gameplay sequence: (Top left) Initial state showing black-marked enemies to defeat and items to collect for rewards. (Top centre) Player spawns and controls allies as additional units. (Top right) Allies collect items while enemies advance each turn. (Bottom left) One ally is defeated by an enemy while simultaneously eliminating an opposing unit. (Bottom centre) Player and remaining ally attempt coordinated attack but are overwhelmed by enemies, resulting in a loss.}
\label{gameA}
\end{figure*}
\begin{figure*}[h!]
\centering
\includegraphics[width=0.9\textwidth]{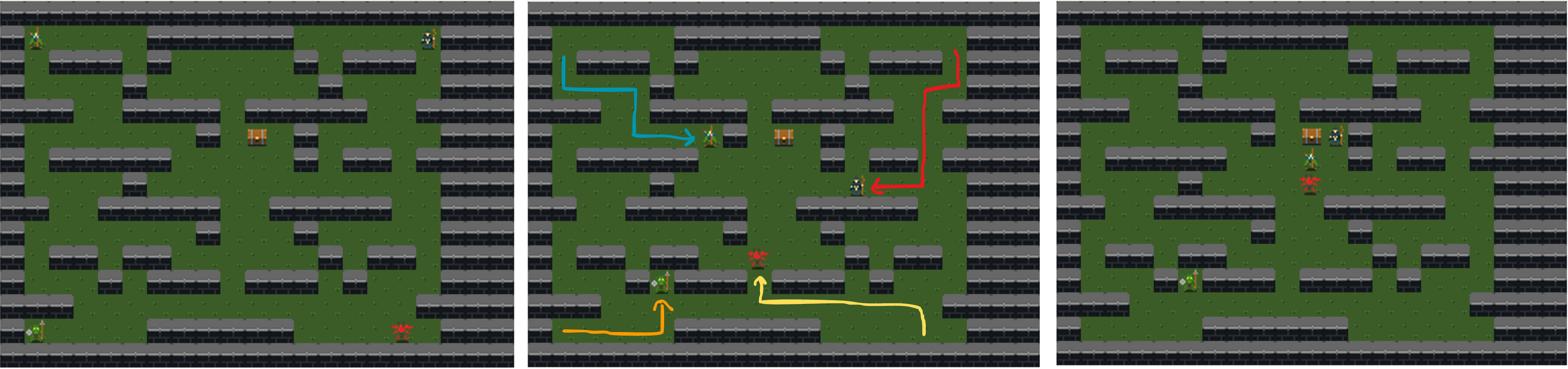} 
\caption{\textit{TreasureHunt} gameplay sequence: (Left) Initial game state showing treasure objective in a capture-the-flag style layout. (Centre) Player spawns at the top-left corner (blue marker) while enemies begins pursuit. (Right) Final state showing close competition between player and red-marked enemy, with victory determined by action processing order. The game features an evolved A* pathfinding algorithm for enemy movement (code in Appendix~\ref{appendixB}).}
\label{gameB}
\end{figure*}

Figures~\ref{gameA} and \ref{gameB} showcase two games generated by \name{}, demonstrating diversity in level layouts, win conditions, and mechanics. \textit{AllyCraft} (Figure~\ref{gameA}) presents a challenging strategic experience where players control both their character and summoned allies, with escalating difficulty requiring versatile tactics. Effective strategies involve summoning allies strategically and eliminating enemies in optimal sequences. This game achieves a Kendall's $\tau$ of 0.8, maintaining clear agent rankings despite low overall rewards, with only minor rank switching between the do-nothing and random agents due to negative scoring.

By contrast, \textit{TreasureHunt} (Figure~\ref{gameB}) exhibits a Kendall's $\tau$ of 0.4, showing significant rank distortion except for the top-performing agent. This lower correlation suggests reduced strategic depth—once players discover the optimal path, the game loses replay value. \textit{AllyCraft}'s higher $\tau$  score correlates with sustained engagement through multiple viable strategies, while \textit{TreasureHunt}'s deterministic solution path limits long-term interest. Both games incorporate sophisticated mechanics, including ally summoning, multi-unit control, and pathfinding algorithms. The complete evolved code for these mechanics is provided in Appendix~\ref{appendixB}.

\subsection{User study}
To determine whether the quantitative metrics correlate with human preferences, we conducted a small comparative user study with 10 participants who evaluated 6 games generated by \name{} across five dimensions: interestingness, novelty, frustration level, fun factor, and ease of understanding.
Table~\ref{userstudy} presents the results alongside each game's Kendall's $\tau$ score for comparison with \name{}'s automated evaluation.

The study compared three pairs of games: \textit{TreasureHunt} versus \textit{HuntBreakout} (capture-the-flag variants where \textit{HuntBreakout} adds wall-breaking mechanics), \textit{AllyCraft} versus \textit{CrystalCavernsCommander} (RPG-style games differing in ally control mechanisms), and \textit{MagneticProwess} versus \textit{HeroHunt} (Sokoban-based games with magnetic pulling and enemy combat mechanics, respectively). See Appendix~\ref{appendixD} for games in the user study.


We observe a general correlation between the total human preference score and \name's calculated Kendall's $\tau$ values. In the first comparison, the $\tau$ difference has a smaller magnitude than the total score difference, yet both favour the same game. The second comparison shows alignment in both magnitude and direction between total score and $\tau$. However, the third comparison reveals opposing trends where the total score contradicts $\tau$, though this discrepancy may reflect the inherent difficulty of aligning automated skill-based metrics with subjective human preferences across diverse game genres.

Treating the total score as a meaningfulness metric---comprising interestingness, novelty, fun factor, ease of understanding, minus frustration---the ``Neither'' votes provide additional insight. These scores (1, 7, and 2 across the three comparisons, respectively) indicate that games in the second comparison are perceived as less meaningful, likely due to excessive complexity. This finding aligns with intuitive game design principles: mini-games benefit from appropriate rather than maximal complexity. While complexity can enhance engagement in full games through progressive difficulty scaling, these mini-game experiences demonstrate reduced effectiveness when sophisticated mechanics overwhelm fundamental gameplay elements.
Finally, qualitative participant feedback consistently highlighted visual limitations, particularly the absence of animations and restricted sprite sets---a known limitation of \name{}'s current implementation. 



\begin{table*}[h]
\centering
\caption{User study results comparing games. Values indicate the number of participants (out of 10) selecting each option. Total score represents the sum of positive metrics minus ``Frustrating''.} \label{userstudy}
\resizebox{\textwidth}{!}{%
\begin{tabular}{@{}lcccccccc@{}}
\toprule
\textbf{Games} & \textbf{Interesting $\uparrow$} & \textbf{Novel $\uparrow$} & \textbf{Frustrating $\downarrow$} & 
\textbf{Fun to play $\uparrow$} & \textbf{Easy to understand $\uparrow$} & \textbf{Total $\uparrow$} & \boldmath{$\tau \uparrow$} \\
\midrule
\textit{TreasureHunt} & 0 & 1 & 3 & 1 & 4 & 3 & 0.4 \\
\textit{HuntBreakout} & 8 & 8 & 5 & 7 & 4 & 22 & 0.5 \\
Both & 1 & 0 & 0 & 1 & 2 & 2 & --- \\
Neither & 1 & 1 & 2 & 1 & 0 & 1 & --- \\
\midrule \midrule
\textit{AllyCraft} & 7 & 6 & 5 & 6 & 3 & 17 & 0.8 \\
\textit{CrystalCavernsCommander} & 2 & 2 & 3 & 3 & 2 & 6 & 0.3 \\
Both & 0 & 1 & 2 & 0 & 1 & 0 & --- \\
Neither & 1 & 1 & 0 & 1 & 4 & 7 & --- \\
\midrule \midrule
\textit{MagneticProwess} & 4 & 4 & 5 & 4 & 3 & 10 & 0.6 \\
\textit{HeroHunt} & 5 & 5 & 2 & 5 & 3 & 16 & 0.3 \\
Both & 0 & 0 & 1 & 0 & 3 & 2 & --- \\
Neither & 1 & 1 & 2 & 1 & 1 & 2 & --- \\
\bottomrule
\end{tabular}}

\end{table*}

\section{Related Work}

The core focus of \name{} is evolving game mechanics to serve as an ideation and prototyping tool for game designers and generate novel games for testing learning algorithms. This research falls under Automatic Game Design (AGD), pioneered by~\citep{nelson2007towards}, who formalized game mechanics through WordNet to generate micro games. \citet{browne2008automatic} and \citet{togelius2008experiment} independently proposed evolutionary approaches to AGD across different domains. The latter introduced learnability as a quality criterion, inspiring various approximations of skill differentiation over the years~\citep{nielsen2015general,khalifa2017general} that influence our current approach. \citet{sumner2024mechanic} introduced Mechanic Maker, a mechanics generation pipeline that used a Domain-Specific Language (DSL) to integrate the mechanic in the game and then created an automated frontend generation as well for the mechanic. Later on, \cite{gonzalez2023mechanic} used Reinforcement Learning to evaluate the generated mechanics in Mechanic Maker 2.0. A very closely related work is Pixie, introduced by \citet{cook2025pixie}, which used Genetic Programming style code evolution to generate mechanics. This work is closely related to \name{} as we also evolve code for the game mechanics.  used playtrace information to identify important game mechanics. \names{} Constrained Importance Through Search (CITS) is related to \citet{green2019automatic} work which uses playtrace information, through comparing different MCTS setups, to find important game mechanics. 

Related concepts include game depth~\citep{lantz2017depth} and formalisms for measuring a game's ability to distinguish among agents~\citep{stephenson2020continuous}.

Non-evolutionary AGD approaches include constraint solvers for mechanics generation~\citep{zook2014automatic} and autoencoders for learning and generating mechanics~\citep{rieder2018using}. Recent work incorporates LLMs into AGD pipelines: ScriptDoctor generates PuzzleScript games~\citep{earle2025scriptdoctor}, while Gavel evolves Ludii games using LLMs and Quality-Diversity algorithms~\citep{todd2024gavel}. Similar approaches have generated 2-player games using XML-based languages~\citep{jorge2023automated}. \name{} distinguishes itself by leveraging the full expressiveness of Python code generation, creating a search space that scales with advancing LLM capabilities.




\name{} also relates to LLM-driven Procedural Content Generation~\citep{togelius2011search, shaker2016procedural, liu2021deep}. Early work included Sokoban level generation using GPT-2/3~\citep{todd2023level}, MarioGPT for Super Mario Bros levels with Novelty Search~\citep{sudhakaran2023mariogpt}, and human-in-the-loop GPT-3 fine-tuning~\citep{nasir2023practical}. Word2World and Word2Minecraft generate 2D and 3D games with fixed mechanics~\citep{nasir2024word2world, huang2025word2minecraft}. \name{} extends this paradigm by generating multiple game aspects, including mechanics and levels.

Finally, \name{} contributes to research on LLMs as evolutionary operators in Quality-Diversity (QD) algorithms like MAP-Elites~\citep{mouret2015illuminating}. This approach has been applied to robot morphology evolution~\citep{lehman2023evolution}, and neural architecture search~\citep{nasir2024llmatic}. Notably, GAVEL~\citep{todd2024gavel} is very similar to our work as it also uses QD algorithm to evolve games, but the focus is on board games.



\section{Limitations}

While \name{} successfully generates novel game mechanics and coherent games with semantically meaningful CITS scores, several limitations warrant future investigation. The system currently modifies game rendering functions without incorporating animations, limiting visual richness. Our experiments used a relatively modest LLM (GPT-4o-mini); stronger models could potentially yield more sophisticated mechanics and improved code quality. The current 2D top-down perspective constrains the search space—extending to 3D environments would significantly expand creative possibilities.

Archive initialisation presents another challenge, as improved seeding strategies could enhance convergence and final quality. Similarly, increasing MCTS iterations during evaluation might produce higher-quality games at the cost of computational resources. Perhaps most significantly, \name{}'s autonomous evolution lacks designer control mechanisms. A controllable variant that accepts design constraints or preferences could better serve as an ideation tool, allowing game developers to guide the search toward specific gameplay goals while maintaining the system's creative discovery capabilities.



\section{Conclusion And Future Directions}

 We present \name{}, a novel system for generating games through mechanic evolution. \name{} combines MAP-Elites, a Quality-Diversity algorithm, with LLM-driven code-level mechanic evolution. The system evaluates mechanics through MCTS, which constructs complete games in each tree node and assesses them using skill-based ranking. We introduce the Constrained Importance Through Search (CITS) score, derived from Shapley values, which quantifies a mechanic's contribution within the actually searched combination space rather than hypothetical alternatives.

Our quantitative results demonstrate \name{}'s high evolvability and progressive improvement across generations through comprehensive ablation studies. Qualitative analysis reveals that games with higher scores exhibit greater strategic depth and complexity, while \name{} consistently produces diverse gaming experiences with sophisticated mechanic interactions.

\name{} offers several promising research directions. As an ideation tool, it could support game designers by suggesting novel mechanic combinations responsive to design constraints. The system's scalability suggests that initialisation with extensive mechanic libraries and extended evolution periods could explore previously undiscovered regions of game design space. The generated games provide rich environments for testing generalisation in reinforcement learning agents~\citep{sutton1999reinforcement}, offering diverse challenges with measurable skill gradients. 

\bibliography{iclr2026_conference}
\bibliographystyle{iclr2026_conference}

\appendix

\section{Prompts}
\label{appendixA}

\subsection{Game Mechanic Generation Prompts}

Prompts provided to \name{} are accompanied by predefined Python code, which is then modified as required. The following methods of a class are defined separately so they can be invoked in individual prompts:

\begin{lstlisting}

init_method = r"""def __init__(self, walkable_tiles,tiles_without_char,  tiles, str_map_without_chars, str_map, interactive_object_tiles, enemy_tiles, render_mode="human"):
    super(GameMechEnv, self).__init__()
    self.map_str_without_chars = str_map_without_chars.strip().split('\n')
    self.map_str = str_map.strip().split('\n')
    self.map = [list(row) for row in self.map_str]
    self.map_without_chars = [list(row) for row in self.map_str_without_chars]
    self.tiles = tiles
    self.tiles_without_char = tiles_without_char
    self.action_space = spaces.Discrete(self.get_action_space())
    self.char_set = {'A': 0, 'B': 1, 'C': 2, 'D': 3, 'O': 4, '@': 5, '#': 6, '&': 7}
    self.char_to_int = lambda c: self.char_set.get(c, 0)
    self.mechanic_to_action = self.get_mechanics_to_action()
    self.done = False
    self.tile_size = 16
    self.char_tile_size = 16
    self.frames = []
    max_width = max(len(row) for row in self.map_str)
    self.observation_space = spaces.Box(
        low=0, 
        high=1,
        shape=(len(self.char_set), len(self.map_str), max_width),  # Use len(self.char_set) for channels
        dtype=np.int32
    )
    
    self.default_walkable_tile = 'A'
    self.render_mode = "rgb_array"
    self.walkable_tiles = walkable_tiles
    self.interactive_object_tiles = interactive_object_tiles
    self.enemy_tiles = enemy_tiles
    self.picked_objects = []
    self.npc_tiles = ["&"]
    self.enemy_tiles = ["#"]
    self.player_health = 100
    self.enemy_health = 100
    self.current_score = 0
    self.map = [list(row) for row in self.map_str]
    self.grid_width = max(len(row) for row in self.map)
    self.grid_height = len(self.map)
    for i, row in enumerate(self.map):
        for j, tile in enumerate(row):
            if tile == '@':
                self.player_position = (i, j)
    
    self.reset()"""

    reset_method = """def reset(self, seed=None):
    self.map = [list(row) for row in self.map_str]
    self.map_without_chars = [list(row) for row in self.map_str_without_chars]
    self.grid_width = max(len(row) for row in self.map)
    self.grid_height = len(self.map)
    for i, row in enumerate(self.map):
        for j, tile in enumerate(row):
            if tile == '@':
                self.player_position = (i, j)
    self.current_tile = self.map_without_chars[self.player_position[0]][self.player_position[1]]  # Set current tile to the player's starting position
    return self.get_state()["map"]"""

    render_method = """def render(self, mode='human'):

    env_img = Image.new('RGBA', (len(self.map[0]) * self.tile_size, len(self.map) * self.tile_size))

    # 1st layer: Default walkable tile
    for i in range(len(self.map)):
        for j in range(len(self.map[0])):
            tile_img = self.tiles[self.default_walkable_tile].resize((self.tile_size, self.tile_size))
            env_img.paste(tile_img, (j * self.tile_size, i * self.tile_size), tile_img)
    
    # 2nd layer: Map without characters
    for i, row in enumerate(self.map_without_chars):
        for j, tile in enumerate(row):
            if tile in self.tiles and tile != self.default_walkable_tile:
                tile_img = self.tiles[tile].resize((self.tile_size, self.tile_size))
                env_img.paste(tile_img, (j * self.tile_size, i * self.tile_size), tile_img)
    
    # 3rd layer: Characters and objects
    for i, row in enumerate(self.map):
        for j, tile in enumerate(row):
            if tile in self.tiles and tile not in self.walkable_tiles:
                if tile.isalpha():
                    tile_img = self.tiles[tile].resize((self.tile_size, self.tile_size))
                else:
                    tile_img = self.tiles[tile].resize((self.char_tile_size, self.char_tile_size))
                    # Center the character in the tile
                    x_offset = (self.tile_size - self.char_tile_size) // 2
                    y_offset = (self.tile_size - self.char_tile_size) // 2
                    env_img.paste(tile_img, (j * self.tile_size + x_offset, i * self.tile_size + y_offset), tile_img)
    
    frame = np.array(env_img.convert('RGB'))
    self.frames.append(frame)
    return frame"""

    get_action_space_method = """def get_action_space(self):
    return 3"""

    get_mechanics_to_action_method = """def get_mechanics_to_action(self):
    return {
        "move_player": 0,  # 0-3 for movement
    }"""

    move_player = """def move_player(self, action):
    moves = {0: (-1, 0), 1: (1, 0), 2: (0, -1), 3: (0, 1)}  # Up, Down, Left, Right
    dx, dy = moves[action]
    new_row = self.player_position[0] + dx
    new_col = self.player_position[1] + dy
    reward = 0
    if 0 <= new_row < len(self.map) and 0 <= new_col < len(self.map[0]):
        new_tile = self.map[new_row][new_col]
        if new_tile in self.walkable_tiles:
            self.update_player_position(new_row, new_col, new_tile)
    return reward"""

    other_methods = r"""def update_player_position(self, new_row, new_col, new_tile):
    # Validate both current and new positions are within bounds
    if not (0 <= new_row < self.grid_height and 0 <= new_col < self.grid_width):
        return
        
    if not (0 <= self.player_position[0] < self.grid_height and 0 <= self.player_position[1] < self.grid_width):
        # If current position is invalid, just set the new position
        self.player_position = (new_row, new_col)
        self.current_tile = new_tile
        self.map[new_row][new_col] = '@'
        return
            
    if new_tile not in self.walkable_tiles:
        return
            
    # Reset the player's previous position to the original tile
    self.map[self.player_position[0]][self.player_position[1]] = self.current_tile
    self.map_without_chars[self.player_position[0]][self.player_position[1]] = self.current_tile
    
    # Update the player's position
    self.player_position = (new_row, new_col)
    self.current_tile = new_tile
    self.map[new_row][new_col] = '@'

def find_player_position(self):
    for i, row in enumerate(self.map):
        for j, tile in enumerate(row):
            if tile == '@':
                return (i, j)
    return None
    
def clone(self):

    new_env = GameMechEnv(
        walkable_tiles=self.walkable_tiles,
        tiles_without_char=self.tiles_without_char,
        tiles=self.tiles,
        str_map_without_chars='\n'.join(self.map_str_without_chars),
        str_map='\n'.join(self.map_str),
        interactive_object_tiles=self.interactive_object_tiles,
        enemy_tiles=self.enemy_tiles
    )
    new_env.map = [row[:] for row in self.map]
    new_env.map_without_chars = [row[:] for row in self.map_without_chars]
    new_env.player_position = self.player_position
    new_env.current_tile = self.current_tile
    new_env.char_to_int = self.char_to_int
    new_env.char_set = self.char_set
    return new_env

def is_terminal(self):
    return self.done"""

    get_state_method = """def get_state(self):
    return {"map": self.map}"""

    step_method = """def step(self, action):
    reward = 0
    self.done = False
    if action < 4:  # Movement actions
        self.move_player(action)
    self.done = reward > 0
    info = {}
    return self.get_state()["map"], reward, self.done, False, info
    """

\end{lstlisting}

The following are the prompts used to edit the methods:

\begin{enumerate}

\item The state, render, and step method prompts are identical, except that the input function is replaced by the function currently in focus:

\begin{lstlisting}
    "Given the following get_state method of the class:\n" + get_state_method + "\n And the following game mechanic:\n" + mechanics + "\n Edit get_state method, if required, for the given game mechanic to work. Do not repeat the game mechanic as a method. Only output whole of the edited get_state method and if not edited, just output 'False'. Do not output anything else."
\end{lstlisting}

\item Adding helper methods:

\begin{lstlisting}
    "Given the following methods: init method of the class:\n" + init_method + "\n All functions already present in the class:\n"+ other_methods + "\n The following game mechanic you just created:\n" + mechanics + "\n Add any new helper methods needed for the " + extract_function_name(mechanics) +" to work, if required. Do not create or update __init__, reset, step, get_state, render, get_action_space, or get_mechanics_to_action methods. Do not repeat the game mechanic as a method or the present methods. Do not add any new variables. Only output the additional new method or methods required, and if not added, just output 'False'. Do not output anything else."
\end{lstlisting}

\item The prompt to add new variables. \lstinline|game\_mech\_class| one string with the whole initial class present in it.

\begin{lstlisting}
    "Given the following class:\n" + game_mech_class + "\n And the following game mechanic:\n" + mechanics + "\n Add any new variables required in the GameMechEnv for the given game mechanic to work. Always add to the existing code. Only output the new variables if they are added in a Python dictionary format and if not added, just output 'False'. The Python dictionary can look like {'var_name_1': init_value, 'var_name_2': init_value}. The keys should be the string of the variable name and the values should be the initial values. The values can never be a new argument to the init method. Do not output anything else."
\end{lstlisting}

\item To generate the action space

\begin{lstlisting}
    "Given the following class:\n" + game_mech_class + "\n Also, the following game mechanic:\n" + mechanics + "\n And the following get_action_space method:\n" + action_space + "\n . Edit the get_action_space method to add more actions, if required, for the given game mechanic to work. Do not repeat the game mechanic as a method. Only output the edited get_action_space method and if not edited, just output 'False'. Do not output anything else."
\end{lstlisting}

\item To get a mapping of mechanics to actions.

\begin{lstlisting}
    "Given the following class:\n" + game_mech_class + "\n Also, the following game mechanic:\n" + mechanics + "\n And the following get_mechanics_to_action method:\n" + mech_to_action + "\n . Edit the get_mechanics_to_action method to add more actions, if required, for the given game mechanic to work. The edition should be "+ extract_function_name(mechanics) +": action_number. Do not repeat the game mechanic as a method. Only output the edited get_mechanics_to_action method and if not edited, just output 'False'. Do not output anything else."
\end{lstlisting}

\item Game mechanics are then tested on a static environment:

\begin{lstlisting}
str_world = """BBBBBBBBBBB
BAAAAAAAAAB
BAAAOAAAAAB
BA#@OAAAAAB
BA#AAAAAAAB
BBBBBBBBBBB"""

str_map_wo_chars = """BBBBBBBBBBB
BAAAAAAAAAB
BAAOOAAAAAB
BAAAOAAAAAB
BAAAAAAAAAB
BBBBBBBBBBB"""

walkables = ['A', 'B']
interactive_object_tiles = ['O']
enemy_tiles = ["#"]
npc_tiles = ["&"]
env_image = dict()

env_image["A"] = Image.open(r"world_tileset_data/td_world_floor_grass_c.png").convert("RGBA")
env_image["B"] = Image.open(r" world_tileset_data/td_world_wall_stone_h_a.png").convert("RGBA")
env_image["C"] = Image.open(r"world_tileset_data/td_world_floor_grass_c.png").convert("RGBA")
env_image["O"] = Image.open(r"world_tileset_data/td_world_chest.png").convert("RGBA")
env_image["@"] = Image.open(r"character_sprite_data/td_monsters_archer_d1.png").convert("RGBA")
env_image["#"] = Image.open(r"character_sprite_data/td_monsters_witch_d1.png").convert("RGBA")
env_image["&"] = Image.open(r"character_sprite_data/td_monsters_goblin_captain_d1.png").convert("RGBA")

env = GameMechEnv(walkable_tiles=walkables, 
        tiles_without_char=str_map_wo_chars, 
        tiles=env_image, 
        str_map_without_chars=str_map_wo_chars, 
        str_map=str_world, 
        interactive_object_tiles=interactive_object_tiles, 
        enemy_tiles=enemy_tiles)
\end{lstlisting}

\end{enumerate}

\subsection{Game Generation Prompts}

All of the above prompts are also used in the game‑generation pipeline. The difference is that we keep track of the previously edited method so that, when the evaluation MCTS tree expands to the next node, the prompt includes methods inherited from earlier nodes. This is necessary because each newly expanded node introduces a new mechanic in addition to all prior mechanics.

After these generations, we generate a \lstinline|make\_game| function. We start with generating a \lstinline|env\_dict\_func|, a function that returns a dictionary mapping tiles to their corresponding path functions. We provide the following paths:

\begin{lstlisting}
    paths_to_tiles = r'''world_tileset_data/td_items_amulet_gold.png,
        world_tileset_data/td_items_gem_ruby.png,
        world_tileset_data/td_world_crate.png,
        world_tileset_data/tg_world_barrel.png,
        world_tileset_data/tg_world_floor_carpet_d.png,
        world_tileset_data/tg_world_floor_moss_e.png,
        world_tileset_data/tg_world_floor_sand_f.png,
        world_tileset_data/tg_world_floor_panel_steel_c.png,
        character_sprite_data/td_monsters_angel_d2.png,
        character_sprite_data/td_monsters_archer_u2.png,
        character_sprite_data/td_monsters_berserker_d1.png,
        character_sprite_data/td_monsters_demon_l1.png'''
\end{lstlisting}

And the whole initial \lstinline|env\_dict\_func|:

\begin{lstlisting}
    env_dict_func = '''def env_dict():
    env_image = dict()
    image_paths = dict()  # New dictionary to store paths
    
    # Define a function to load image and store path
    def load_image(char, path):
        env_image[char] = Image.open(path).convert("RGBA")
        image_paths[char] = path  # Store the path
    
    # Load all images
    base_path = r"/mnt/lustre/users/mnasir/gmd"
    load_image("A", f"{base_path}/world_tileset_data/td_world_floor_grass_c.png")
    load_image("B", f"{base_path}/world_tileset_data/td_world_wall_stone_h_a.png")
    load_image("X", f"{base_path}/world_tileset_data/td_world_floor_grass_c.png")
    load_image("O", f"{base_path}/world_tileset_data/td_world_chest.png")
    load_image("I", f"{base_path}/world_tileset_data/td_world_chest.png")
    load_image("C", f"{base_path}/world_tileset_data/td_world_chest.png")
    load_image("@", f"{base_path}/character_sprite_data/td_monsters_archer_d1.png")
    load_image("#", f"{base_path}/character_sprite_data/td_monsters_witch_d1.png")
    load_image("&", f"{base_path}/character_sprite_data/td_monsters_goblin_captain_d1.png")

    # Here you add any other tiles that are needed for the game in the same format
    
    return env_image, image_paths '''
\end{lstlisting}

Therefore, the prompt for the \lstinline|env\_dict\_func| generation:

\begin{lstlisting}
    "Given the game mechanics:\n" + mechanics[0] + "\nChange the following env_dict function to cater for the mechanics, if required.:\n" + env_dict_func + " Use the same paths for the same type of tiles that are being added in env_image, or use the following paths:\n"+ paths_to_tiles + "\n You don't have to use the paths provided if not needed. Strictly use the same paths. Do not use any other paths. Only add in env_image if needed. Return the full env_dict function."
\end{lstlisting}

Then we create a 2D character map through the prompt:

\begin{lstlisting}
    "Given the game mechanics:\n" + mechanics[0] + "\n The env_dict function, which has the paths for the tiles:\n" + env_dict_func_changed['choices'][0]['message']['content'] + "\nChange the following str_world in the str_map function to cater for the mechanics, if required.:\n" + str_map_func + "\n 'str_world' is the string that represents the 2D game map. Change it if only needed. It must always have 1 and only 1 `@` character, which represents the player. Return the full str_map function."
\end{lstlisting}

where the \lstinline|str\_map\_func| is:

\begin{lstlisting}
    str_map_func = '''def str_map():

        str_world = """AAAAAAAAAAAAAAAAAA
    AAAAAAAAAAAAAAAAAA
    AAA@OAXAAAAAAAAAAA
    AAAAAAICAAAAAAAAAA
    AAA#AAAAAAAAAAAAAA
    AAAAAAAAAAAAAAAAAA"""
    
        return str_world'''
\end{lstlisting}

Lastly, in the \lstinline|make\_game| function we generate the \lstinline|important\_tiles\_func| through the prompt:

\begin{lstlisting}
    "Given the game mechanics:\n" + mechanics[0] + "\n And the env_dict function, which has the paths for the tiles:\n" + env_dict_func_changed['choices'][0]['message']['content'] + "\n And the str_map function, which has the string representation of the game map:\n" + str_map_func_changed['choices'][0]['message']['content'] + "\n Change the following important_tiles function to cater for the mechanics, if required.:\n" + important_tiles_func + " Return the full important_tiles function. Return all the tile types mentioned in the return statement of the function. Return empty list if the tile type is not needed."
\end{lstlisting}

where the initial \lstinline|important\_tiles\_func| is:

\begin{lstlisting}
    important_tiles_func = '''def important_tiles():
        walkables = ['A']  # Walkable tiles
        non_walkables = ['B']  # Non-walkable tiles
        interactive_object_tiles = ['O', 'I', 'C']  # Interactive objects (e.g., chests)
        collectible_tiles = []  # Can add collectible tiles if needed
        npc_tiles = []  # Assume there are no NPCs represented in the current setup
        player_tile = ['@']  # Player tile
        enemy_tiles = ['#', '&']  # Enemy tiles
        extra_tiles = [] # any other type of tiles for the game goes here
        return walkables, non_walkables, interactive_object_tiles, collectible_tiles, npc_tiles, player_tile, enemy_tiles, extra_tiles'''

\end{lstlisting}

For game‑mechanic generation, the terminal method is fixed, since we test each mechanic in isolation to verify that the MCTS agent can reach the end. In the subsequent game‑generation stage, however, the terminal method may vary, as the win condition can change substantially.

\begin{lstlisting}
    "Given the game mechanics:\n" + get_games_scores.latest_methods['mechanic'] + "\n" + mechanics[0] + "\n and the init function:\n" + get_games_scores.latest_methods['init'] + "\n We want to train an agent to play a game that uses these mechanics. The layout of the game is the str_world in the following function:\n"+ get_games_scores.latest_methods['str_world'] +"\n and the following function describes what the tiles mean:\n"+get_games_scores.latest_methods['tiles']+"\nThe following line describes the situation of the win condition of the game:\n"+ "'"+line_response['choices'][0]['message']['content']+"'" +"\nWrite one method of the class which which wraps win condition in it and tells the agent when the game will end. It must focus on the mechanics in the game provided to you. All the mechanics should be used to fulfill the win condition. The name of the method should be is_terminal. Method should only return one boolean variable. Only return the method and nothing else."
\end{lstlisting}

Here a \lstinline|line\_response| is a win condition generated through the following prompt:

\begin{lstlisting}
    "Given the game mechanics:\n" + get_games_scores.latest_methods['mechanic'] + "\n" + mechanics[0] + "\n write one line that describes the win condition for the game that will use these mechanics. "
\end{lstlisting}

The name of the game is generated through:

\begin{lstlisting}
    "Given the game mechanics:\n" + get_games_scores.latest_methods['mechanic'] + "\n" + mechanics[0] + "\nThe win condition for the game in is_terminal method:\n " + is_terminal_response['choices'][0]['message']['content'] + "\n And the line that explains the win condition:\n " + line_response['choices'][0]['message']['content'] + "\n Give the game a short name that describes the game well. Only strictly output the name and nothing else. Should not have any special characters in the name. Do not highlight the name."
\end{lstlisting}

\section{Generated Game Mechanics} 
\label{appendixB}

\subsection{Mechanics In Figure 6}

Following is the mechanic and helper functions for the game in Figure 6:

\begin{lstlisting}
def spawn_unit(self):
    """Spawn a unit at an adjacent empty position"""
    reward = 0
    
    if len(self.units) >= self.max_units:
        return 0  # No penalty, just no reward
        
    player_row, player_col = self.player_position
    adjacency_offsets = [(0, -1), (0, 1), (-1, 0), (1, 0)]  #Left, Right, Up, Down
    
    # Try to find an empty adjacent position
    for dx, dy in adjacency_offsets:
        new_row = player_row + dx
        new_col = player_col + dy
        if (0 <= new_row < len(self.map) and 0 <= new_col <len(self.map[0])):
            # Check the base tile type (without characters)
            base_tile = self.map_without_chars[new_row][new_col]
            current_tile = self.map[new_row][new_col]
            
            # Check if position is suitable for unit spawning
            if (base_tile in self.walkable_tiles and 
                (new_row, new_col) not in self.units and
                (new_row, new_col) != self.player_position and
                current_tile not in self.enemy_tiles and
                current_tile in self.walkable_tiles):  # Current tile should also be walkable
                
                # Spawn unit here
                unit_pos = (new_row, new_col)
                self.units.append(unit_pos)
                self.unit_health[unit_pos] = 100  # Initialize unit health
                self.map[new_row][new_col] = self.unit_symbol
                reward = 1  # Reward for successful spawning
                break
    
    return reward

#-------------------------

    def heal_unit(self):
        """Heal the selected unit"""
        reward = 0
        
        if not self.units or self.selected_unit >= len(self.units):
            return -1  # Penalty for invalid unit selection
            
        unit_pos = self.units[self.selected_unit]
        
        if unit_pos in self.unit_health:
            old_health = self.unit_health[unit_pos]
            self.unit_health[unit_pos] = min(100, self.unit_health[unit_pos] + 30)  # Heal 30 HP, max 100
            
            if old_health < self.unit_health[unit_pos]:
                heal_amount = self.unit_health[unit_pos] - old_health
                print(f"Unit at {unit_pos} healed for {heal_amount} HP! Health: {self.unit_health[unit_pos]}")
                reward = 1  # Small reward for healing
            else:
                print(f"Unit at {unit_pos} is already at full health!")
                reward = -1  # Penalty for unnecessary healing
        
        return reward

#-------------------------

def player_attack(self):
    """Execute primary attack on adjacent targets"""
    reward = 0
    player_row, player_col = self.player_position
    adjacency_offsets = [(0, -1), (0, 1), (-1, 0), (1, 0)]
    
    enemies_attacked = 0
    for dx, dy in adjacency_offsets:
        attack_row = player_row + dx
        attack_col = player_col + dy
        attack_pos = (attack_row, attack_col)
        
        # Find enemy at this position
        for enemy in self.enemies:
            if enemy['pos'] == attack_pos:
                damage = 25  # Player damage
                enemy['health'] -= damage
                enemies_attacked += 1
                print(f"Player attacks {enemy['type']} for {damage} damage! Enemy health: {enemy['health']}")
                
                if enemy['health'] <= 0:
                    print(f"{enemy['type']} defeated!")
                    reward += 10  # Reward for defeating enemy
                else:
                    reward += 2  # Small reward for successful attack
    
    # Small penalty if no enemies to attack
    if enemies_attacked == 0:
        reward = -1
        
    return reward

#-------------------------

def move_enemy_toward_target(self, enemy, target_pos):
    """Move enemy one step toward target"""
    reward = 0
    enemy_pos = enemy['pos']
    enemy_row, enemy_col = enemy_pos
    target_row, target_col = target_pos
    
    # Calculate direction to move
    row_diff = target_row - enemy_row
    col_diff = target_col - enemy_col
    
    # Choose move direction (simple pathfinding)
    move_row, move_col = 0, 0
    if abs(row_diff) > abs(col_diff):
        move_row = 1 if row_diff > 0 else -1
    else:
        move_col = 1 if col_diff > 0 else -1
    
    new_row = enemy_row + move_row
    new_col = enemy_col + move_col
    
    # Check if move is valid
    if self._is_valid_enemy_move(enemy_pos, (new_row, new_col)):
        self._execute_enemy_move(enemy, (new_row, new_col))
    
    return reward

#-------------------------

 def confuse_and_teleport_enemies(self):
    """Apply area effect that disrupts enemy positioning"""
    reward = 0
    # Identify all enemy positions and create a list of positions
    enemy_positions = []
    for row in range(len(self.map)):
        for col in range(len(self.map[0])):
            if self.map[row][col] in self.enemy_tiles:  # Useactual enemy tiles from the map
                enemy_positions.append((row, col))
    # If there are enemies on the map, confuse and possibly teleport them
    if enemy_positions:
        enemies_confused = 0
        for enemy_row, enemy_col in enemy_positions:
            # Randomly choose a direction to confuse the enemy
            direction = random.choice(['up', 'down', 'left', 'right'])
            teleport_possible = False
            # Determine the new position for confusion
            new_enemy_row, new_enemy_col = enemy_row, enemy_col
            if direction == 'up' and enemy_row > 0:
                new_enemy_row -= 1
            elif direction == 'down' and enemy_row < len(self.map) - 1:
                new_enemy_row += 1
            elif direction == 'left' and enemy_col > 0:
                new_enemy_col -= 1
            elif direction == 'right' and enemy_col < len(self.map[0]) - 1:
                new_enemy_col += 1
            # Instead of actually moving enemies, just count confusion attempts
            distance_to_player = abs(enemy_row -self.player_position[0]) + abs(enemy_col -self.player_position[1])
            if distance_to_player <= 2:  # If enemy is within close range
                enemies_confused += 1
        
        # Only give reward if multiple enemies were confused
        if enemies_confused >= 2:
            reward = 1
    return reward   

#-------------------------

def activate_and_combine_resources(self):
    """Activates resource gathering and environmental interactionability."""
    reward = 0
    adjacency_offsets = [(0, -1), (0, 1), (-1, 0), (1, 0)]  # Up, Down, Left, Right
    resource_tiles = ['R', 'F', 'W']  # R = Rock, F = Food, W = Wood
    
    # Count adjacent resources and interactive objects
    adjacent_resources = 0
    adjacent_objects = 0
    
    # Check for adjacent resources
    for dx, dy in adjacency_offsets:
        new_row = self.player_position[0] + dx
        new_col = self.player_position[1] + dy
        if 0 <= new_row < len(self.map) and 0 <= new_col <len(self.map[0]):
            adjacent_tile = self.map[new_row][new_col]
            if adjacent_tile in resource_tiles:
                adjacent_resources += 1
                    
    # Check for interactive objects nearby
    for dx, dy in adjacency_offsets:
        new_row = self.player_position[0] + dx
        new_col = self.player_position[1] + dy
        if 0 <= new_row < len(self.map) and 0 <= new_col <len(self.map[0]):
            adjacent_tile = self.map[new_row][new_col]
            if adjacent_tile in self.interactive_object_tiles:
                adjacent_objects += 1
                
    # Only give reward for significant resource/object combinations
    if adjacent_resources >= 2 and adjacent_objects >= 1:
        reward = 5  # Reward only for optimal positioning
                
    return reward
\end{lstlisting}

\subsection{Mechanics In Figure 7}

The following is the mechanic and helper functions for the game in Figure 7:

\begin{lstlisting}
def strategic_enemy_movement(self):
    """Process all enemy actions for this turn using A*path finding"""
    import heapq
    reward = 0
    
    def heuristic(pos, goal):
        return abs(pos[0] - goal[0]) + abs(pos[1] - goal[1])
    
    for enemy in self.enemies[:]:
        if enemy['finished']:
            continue
            
        if self.turn_counter - enemy['last_move_turn'] >= self.enemy_move_cooldown:
            enemy['last_move_turn'] = self.turn_counter
            
            # A* pathfinding to find next best move
            start = enemy['pos']
            goal = self.goal_position
            
            # Priority queue: (f_score, g_score, position, path)
            open_set = [(heuristic(start, goal), 0, start,[start])]
            closed_set = set()
                
            directions = [(-1, 0), (1, 0), (0, -1), (0, 1)]
            max_iterations = 50  # Limit search to prevent lag
            iterations = 0
            optimal_move = None
            
            while open_set and iterations < max_iterations:
                iterations += 1
                f_score, g_score, current, path =heapq.heappop(open_set)
                
                if current in closed_set:
                    continue
                    
                if current == goal:
                    # Return the next move in the optimal path
                    optimal_move = path[1] if len(path) > 1 elseNone
                    break
                
                closed_set.add(current)
                
                for dx, dy in directions:
                    neighbor = (current[0] + dx, current[1] + dy)
                    
                    if (neighbor in closed_set or 
                        notself._is_valid_position_for_pathfinding(neighbor)):
                        continue
                    
                    new_g_score = g_score + 1
                    new_f_score = new_g_score +heuristic(neighbor, goal)
                    new_path = path + [neighbor]
                    
                    heapq.heappush(open_set, (new_f_score,new_g_score, neighbor, new_path))
            
            # Execute move if valid
            if optimal_move and self._is_valid_enemy_move(enemy['pos'], optimal_move):
                self._execute_enemy_move(enemy, optimal_move)
            
            # Check if enemy reached goal
            if enemy['pos'] == self.goal_position:
                if not enemy['finished'] and notself.game_finished:
                    enemy['finished'] = True
                    self.game_finished = True
                self.completion_order.append(f"enemy_{enemy['type']}")
                    print(f"Enemy {enemy['type']} reached thegoal first! ENEMY WINS!")
                    reward -= 100  # Player loses big when enemy wins
                        
    return reward

#-------------------------

def _is_valid_position_for_pathfinding(self, pos):
    """Check if position is valid for pathfinding (allows temporary occupation)"""
    row, col = pos
    if not (0 <= row < len(self.map) and 0 <= col < len(self.map[0])):
        return False
    
    tile = self.map[row][col]
    # Allow movement through walkable tiles and goal
    return tile in self.walkable_tiles or tile == 'F'

#-------------------------

def _is_valid_enemy_move(self, current_pos, new_pos):
    """Check if enemy move is valid"""
    new_row, new_col = new_pos
    if not (0 <= new_row < len(self.map) and 0 <= new_col < len(self.
        map[0])):
        return False
        
    current_tile = self.map[new_row][new_col]
    
    # Can move to walkable tiles or flag
    if current_tile not in self.walkable_tiles and current_tile != 'F':
        return False
        
    # Cannot move to position occupied by player or other enemies
    if new_pos == self.player_position:
        return False
        
    for other_enemy in self.enemies:
        if other_enemy['pos'] == new_pos:
            return False
            
    return True
\end{lstlisting}

\section{Quality-Diversity}
\label{appendixC}

\subsection{Initial Mechanics}

Here we will mention the aspects of the quality-diversity (QD) algorithm that would help in reproducibility, and were not mentioned in the main paper. The following are the initial mechanics used to initialise the QD algorithm:

\begin{lstlisting}
    mech_1 = """\ndef move_player(self, action):
    moves = {0: (-1, 0), 1: (1, 0), 2: (0, -1), 3: (0, 1)}  # Up, Down, Left, Right
    dx, dy = moves[action]
    new_row = self.player_position[0] + dx
    new_col = self.player_position[1] + dy
    reward = 0
    if 0 <= new_row < len(self.map) and 0 <= new_col < len(self.map[0]):
        new_tile = self.map[new_row][new_col]
        if new_tile in self.walkable_tiles:
            self.update_player_position(new_row, new_col, new_tile)
    return reward"""

#------------------------------- 

mech_2 = """\ndef pick_object(self):
    reward = 0
    # Check adjacent tiles for interactive objects and pick them if present
    adjacent_positions = [(0, -1), (0, 1), (-1, 0), (1, 0)]  # Up, Down, Left, Right
    for dx, dy in adjacent_positions:
        row, col = self.player_position  # player_position is in (row, col) format
        new_row = row + dx
        new_col = col + dy
        if 0 <= new_row < len(self.map) and 0 <= new_col < len(self.map[0]):
            target_tile = self.map[new_row][new_col]
            if target_tile in self.interactive_object_tiles:
                self.map[new_row][new_col] = self.default_walkable_tile 
                reward = 1
                break  # Exit after picking up one object
    return reward"""

#-------------------------------

mech_3 = """\ndef hit_enemy(self):
    reward = 0
    # Check adjacent tiles for enemies and hit them if present
    adjacent_positions = [(0, -1), (0, 1), (-1, 0), (1, 0)]  # Up, Down, Left, Right
    for dx, dy in adjacent_positions:
        row, col = self.player_position  # player_position is in (row, col) format
        new_row = row + dx
        new_col = col + dy
        if 0 <= new_row < len(self.map) and 0 <= new_col < len(self.map[0]):  # Check grid bounds
            target_tile = self.map[new_row][new_col]
            if target_tile in self.enemy_tiles:
                self.map[new_row][new_col] = self.default_walkable_tile
                reward = 1
                break  # Exit after hitting one enemy
    return reward"""

#------------------------------- 

mech_4 = """\ndef teleport_player(self):
    # Find all walkable tiles that are not adjacent to the player
    non_adjacent_walkable_positions = []
    adjacency_offsets = [(0, -1), (0, 1), (-1, 0), (1, 0)]  # Up, Down, Left, Right
    reward = 0
    # Search the map for walkable and non-adjacent tiles
    for row in range(len(self.map)):
        for col in range(len(self.map[0])):
            if self.map[row][col] in self.walkable_tiles:
                is_adjacent = False
                for dx, dy in adjacency_offsets:
                    if (row == self.player_position[0] + dx) and (col == self.player_position[1] + dy):
                        is_adjacent = True
                        break
                if not is_adjacent:
                    non_adjacent_walkable_positions.append((row, col))
    # Teleport the player to a random walkable, non-adjacent position
    if non_adjacent_walkable_positions:
        new_position = random.choice(non_adjacent_walkable_positions)
        self.update_player_position(new_position[0], new_position[1], self.map[new_position[0]][new_position[1]])
        reward += 1
    return reward"""

#------------------------------- 

mech_5 = """\ndef swap_positions(self):
    # Find all enemy positions on the map
    enemy_positions = []
    reward = 0
    for row in range(len(self.map)):
        for col in range(len(self.map[0])):
            if self.map[row][col] in self.enemy_tiles:
                enemy_positions.append((row, col))
    # If there are enemies, randomly swap the player's position with an enemy's position
    if enemy_positions:
        swap_with = random.choice(enemy_positions)
        enemy_row, enemy_col = swap_with
        player_row, player_col = self.player_position
        # Swap positions of player and enemy on the map
        self.map[player_row][player_col], self.map[enemy_row][enemy_col] = self.map[enemy_row][enemy_col], self.map[player_row][player_col]
        # Update the player's position to the swapped position
        self.player_position = (enemy_row, enemy_col)
        # Optional: Output the result of the swap
        reward += 1
    return reward"""

#------------------------------- 

mech_6 = """\ndef push_object(self):
    reward = 0
    adjacent_positions = [(0, -1), (0, 1), (-1, 0), (1, 0)]  # Up, Down, Left, Right
    for dy, dx in adjacent_positions:  # Swapped to dy, dx to match map indexing
        y, x = self.player_position  # Player position is in (row, col) format
        new_y, new_x = y + dy, x + dx
        if 0 <= new_y < len(self.map) and 0 <= new_x < len(self.map[0]):  # Check bounds
            target_tile = self.map[new_y][new_x]
            if target_tile in self.interactive_object_tiles:
                push_y, push_x = new_y + dy, new_x + dx  # Push in same direction
                if 0 <= push_y < len(self.map) and 0 <= push_x < len(self.map[0]):
                    if self.map[push_y][push_x] in self.walkable_tiles:
                        self.map[push_y][push_x] = target_tile
                        self.map[new_y][new_x] = self.default_walkable_tile
                        reward = 1
                        break
    return reward"""

#------------------------------- 

mech_7 = """\ndef jump_player(self):
    reward = 0
    # Define possible jump directions
    jump_directions = [(0, -2), (0, 2), (-2, 0), (2, 0)]  # Up, Down, Left, Right (2 tiles)
    for dx, dy in jump_directions:
        row, col = self.player_position  # player_position is in (row, col) format
        mid_row, mid_col = row + dx // 2, col + dy // 2  # Middle tile (jumped over)
        new_row, new_col = row + dx, col + dy  # Landing tile
        # Check if the jump is within bounds
        if 0 <= new_row < len(self.map) and 0 <= new_col < len(self.map[0]):
            target_tile = self.map[new_row][new_col]
            # Check if the landing tile is walkable
            if target_tile in self.walkable_tiles:
                # Perform the jump
                self.update_player_position(new_row, new_col, target_tile)
                reward = 1
                break  # Exit after a successful jump
    return reward"""

#------------------------------- 

mech_8 = """\ndef drop_object(self):
    reward = 0
    # Check adjacent tiles for empty walkable space
    adjacent_positions = [(0, -1), (0, 1), (-1, 0), (1, 0)]  # Up, Down, Left, Right
    for dx, dy in adjacent_positions:
        row, col = self.player_position
        new_row = row + dx
        new_col = col + dy
        # Check if position is within bounds and walkable
        if 0 <= new_row < len(self.map) and 0 <= new_col < len(self.map[0]):
            if self.map[new_row][new_col] in self.walkable_tiles:
                # Place an interactive object
                self.map[new_row][new_col] = self.interactive_object_tiles[0]  # Using first interactive object tile
                reward = 1
                break  # Exit after dropping one object        
    return reward"""

mech_9 = """\ndef enemy_move(self):
    reward = 0
    # Find all enemy positions with "#" tile on the map
    enemy_positions = []
    for row in range(len(self.map)):
        for col in range(len(self.map[0])):
            if self.map[row][col] == "#":
                enemy_positions.append((row, col))
    
    # If there are enemies, move one randomly
    if enemy_positions:
        # Pick a random enemy to move
        enemy_row, enemy_col = random.choice(enemy_positions)
        
        # Define possible move directions (same as player)
        moves = {0: (-1, 0), 1: (1, 0), 2: (0, -1), 3: (0, 1)}  # Up, Down, Left, Right
        
        # Try each direction randomly until we find a valid move
        directions = list(moves.keys())
        random.shuffle(directions)
        
        for action in directions:
            dx, dy = moves[action]
            new_row = enemy_row + dx
            new_col = enemy_col + dy
            
            # Check if the new position is valid
            if 0 <= new_row < len(self.map) and 0 <= new_col < len(self.map[0]):
                new_tile = self.map[new_row][new_col]
                if new_tile in self.walkable_tiles:
                    # Move the enemy
                    self.map[enemy_row][enemy_col] = self.default_walkable_tile
                    self.map[new_row][new_col] = "#"
                    break  # Exit after successful move
    
    return reward"""

mech_10 = """\ndef enemy_hit(self):
    reward = 0
    # Find all enemy positions with "#" tile on the map
    enemy_positions = []
    for row in range(len(self.map)):
        for col in range(len(self.map[0])):
            if self.map[row][col] == "#":
                enemy_positions.append((row, col))
    
    # Check if any enemy is adjacent to the player and can hit
    player_row, player_col = self.player_position
    adjacent_positions = [(0, -1), (0, 1), (-1, 0), (1, 0)]  # Up, Down, Left, Right
    
    for enemy_row, enemy_col in enemy_positions:
        # Check if this enemy is adjacent to the player
        for dx, dy in adjacent_positions:
            check_row = enemy_row + dx
            check_col = enemy_col + dy
            # If the adjacent position matches the player's position
            if check_row == player_row and check_col == player_col:
                # Enemy hits the player
                reward = -1  # Negative reward for player getting hit
                break  # Exit after first hit (one enemy hitting is enough)
        if reward != 0:  # If a hit occurred, stop checking other enemies
            break
    return reward"""
\end{lstlisting}

\subsection{Game Mechanics Types}

We specify the types of mechanics that \name{} uses to compute similarity scores for the Quality–Diversity archive. For each mechanic, we list its category followed by the keywords used to determine similarity.

 \begin{itemize}
     \item \textbf{Movement}: move, walk, run, jump, fly, teleport, dash, swim, climb, crouch, sprint
     \item \textbf{Interaction}: pick, use, interact, open, close, talk, trade, craft, activate, push, pull
     \item \textbf{Combat}: attack, fight, hit, shoot, defend, block, dodge, cast, spell, heal, damage
     \item \textbf{Progression}: level, upgrade, unlock, improve, evolve, progress, achieve, complete, quest, mission
     \item \textbf{Environment}: weather, day, night, season, climate, destroy, build, terraform, grow, plant
\item \textbf{Puzzle}: solve, puzzle, riddle, match, connect, arrange, decode, decipher, logic, pattern 
\item \textbf{Resource Management}: collect, gather, manage, inventory, store, spend, earn, balance, allocate, distribute 
\item \textbf{Exploration}: explore, discover, map, reveal, uncover, navigate, search, investigate, scout, survey 
\item \textbf{Time Manipulation}: time, slow, fast, rewind, forward, pause, resume, loop, cycle, sequence         

 \end{itemize}

\subsection{Prompts For Evolutionary Operators}

The following are the prompts for the evolutionaru operators:

\begin{enumerate}

\item \textbf{Mutation:}

\begin{lstlisting}
    "Create a new game mechanic from the given mechanic that extends its features:\n" + solution[0] + "\n Do not make any assumptions, if you want to add a new variable or a new function, you should do it within the game mechanic method. The mechanic must return a reward, which is an integer. If a tile is being assumed then it should be defined as a single capital alphabet character and not a word. If a player is being assumed then it should be '@' tile. Remember that the game mechanic function should only take 'self' as parameter. Only output the new game mechanic as Python function, nothing else."
\end{lstlisting}

\item \textbf{Diversity Mutation:}

\begin{lstlisting}
    "Create a new game mechanic that is different, in terms of behavior of mechanics, from the ones provided:\n" + solution[0] + "\n Do not make any assumptions, if you want to add a new variable or a new funciton, you should do it within the game mechanic method. The mechanic must return a reward, which is an integer. If a tile is being assumed then it should be defined as a single capital alphabet character and not a word. If a player is being assumed then it should be '@' tile. Remember that the game mechanic function should only take 'self' as parameter. Only output the new game mechanic as Python function, nothing else."
\end{lstlisting}

\item \textbf{Compatibility Mutation:}

\begin{lstlisting}
    "Create a new game mechanic that will make the game better when combined with the following game mechanics:\n" + solution + "\n Do not make any assumptions, if you want to add a new variable or a new funciton, you should do it within the game mechanic method. The mechanic must return a reward, which is an integer. If a tile is being assumed then it should be defined as a single capital alphabet character and not a word. If a player is being assumed then it should be '@' tile. Remember that the game mechanic function should only take 'self' as parameter. The name of the mechanic should be coherent with the behaviour of it. Only output the new game mechanic as Python function, nothing else."
    
\end{lstlisting}

\item \textbf{Crossover:}

\begin{lstlisting}
    "Create a new game mechanic that combines the features of the given two mechanics to create a new game mechanic that combines the behavior of the both of them:\n" + solution + "\n Do not make any assumptions, if you want to add a new variable or a new method, you should do it within the function. The mechanic must return a reward, which is an integer. If a tile is being assumed then it should be defined as a single capital alphabet character and not a word. If a player is being assumed then it should be '@' tile. Remember that the game mechanic function should only take 'self' as parameter. The name of the mechanic should be coherent with the behaviour of it. Only output the new game mechanic as Python function, nothing else."
    
\end{lstlisting}

\end{enumerate}

\section{Games}
\label{appendixD}
Play games in the user study by following the links:

\begin{enumerate}
    \item TreasureHunt:\url{https://mortar-x3p7.onrender.com/games/TreasureHunt}
    \item HeroBreakout:\url{https://mortar-x3p7.onrender.com/games/HuntBreakout}
    \item AllyCraft:\url{https://mortar-x3p7.onrender.com/games/AllyCraft}
    \item CrystalCavernsCommander:\url{https://mortar-x3p7.onrender.com/games/Crystal_Caverns_Commander}
    \item MagneticProwess:\url{https://mortar-x3p7.onrender.com/games/MagneticProwess}
    \item HeroHunt:\url{https://mortar-x3p7.onrender.com/games/HeroHunt}
\end{enumerate}

\end{document}

%% file: math_commands.tex
\usepackage{amsmath,amsfonts,bm}

\def\eqref#1{equation~\ref{#1}}

\def\1{\bm{1}}

\DeclareMathAlphabet{\mathsfit}{\encodingdefault}{\sfdefault}{m}{sl}
\SetMathAlphabet{\mathsfit}{bold}{\encodingdefault}{\sfdefault}{bx}{n}

